\documentclass[10pt,twocolumn,letterpaper]{article}

\usepackage{cvpr}
\usepackage{times}
\usepackage{epsfig}
\usepackage{graphicx}
\usepackage{amsmath}
\usepackage{amssymb}

\usepackage{color}
\usepackage{amsmath}
\usepackage{amssymb}
\usepackage{multirow, varwidth}
\usepackage{verbatim}
\makeatletter
\newif\if@restonecol
\makeatother

\usepackage[ruled,vlined]{algorithm2e}
\usepackage{xr}
\usepackage[amsmath,thmmarks]{ntheorem}

\theorembodyfont{\normalfont}
\theoremseparator{.}

\theoremstyle{nonumberplain}

\newcommand{\argmax}{\operatornamewithlimits{argmax}}

\DeclareRobustCommand\onedot{\futurelet\@let@token\@onedot}
\def\onedot{.\@\xspace}

\def\eg{\emph{e.g}\onedot} 
\def\ie{\emph{i.e}\onedot}

\def\etal{\emph{et al}\onedot}

\usepackage{array}
\usepackage{gensymb}
\usepackage{color}
\usepackage{tabulary}

\usepackage[breaklinks=true,bookmarks=false]{hyperref}

\cvprfinalcopy 

\def\cvprPaperID{619} 

\ifcvprfinal\fi
\begin{document}

\title{A Memory Network Approach for Story-based \\Temporal Summarization of 360\degree~Videos}

\author{Sangho Lee, Jinyoung Sung, Youngjae Yu, Gunhee Kim\\
Seoul National University\\
{\tt\small sangho.lee@vision.snu.ac.kr, jysung710@gmail.com, yj.yu@vision.snu.ac.kr, gunhee@snu.ac.kr} \\
\tt\small \url{http://vision.snu.ac.kr/projects/pfmn}
}

\maketitle
\thispagestyle{empty}

\begin{abstract}
We address the problem of story-based temporal summarization of long 360\degree~videos. 
We propose a novel memory network model named \textit{Past-Future Memory Network} (PFMN), in which
we first compute the scores of 81 normal field of view (NFOV) region proposals cropped from the input 360\degree~video, 
and then recover a latent, collective summary using the network with two external memories that store the embeddings of previously selected subshots and future candidate subshots. 
Our major contributions are two-fold. 
First, our work is the first to address story-based temporal summarization of 360\degree~videos.
Second, our model is the first attempt to leverage memory networks for video summarization tasks.
For evaluation, we perform three sets of experiments. 
First, we investigate the view selection capability of our model on the Pano2Vid dataset~\cite{su2016accv}. 
Second, we evaluate the temporal summarization with a newly collected 360\degree~video dataset.
Finally, we experiment our model's performance in another domain, with image-based storytelling VIST dataset~\cite{huang2016naacl}. 
We verify that our model achieves state-of-the-art performance on all the tasks. 
\end{abstract}


\begin{figure}[t]
\centering
\includegraphics[trim=0.10cm 0.2cm 0cm 0.1cm,clip,width=0.47\textwidth]{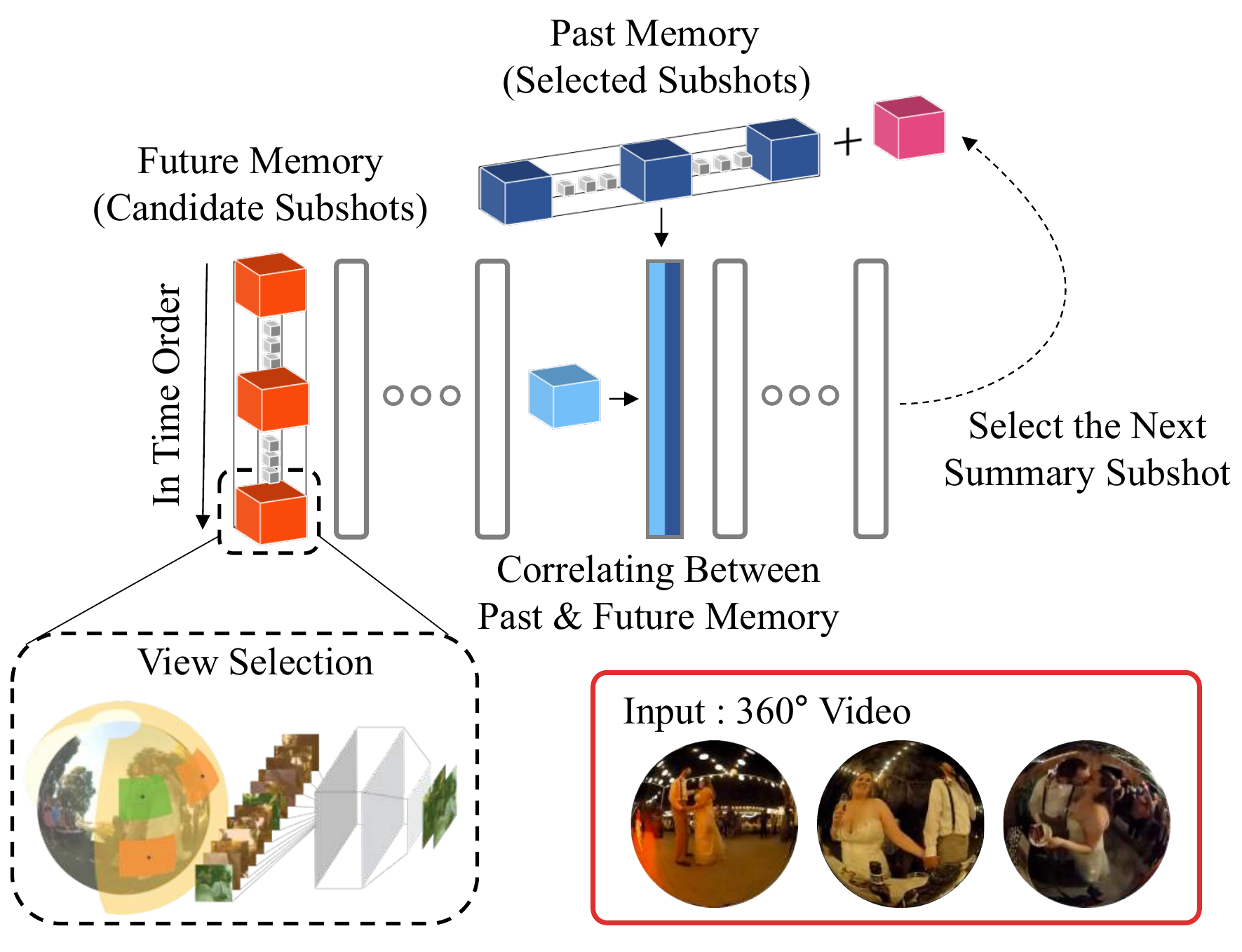}
\caption{The intuition of the proposed PFMN model for temporal summarization of 360\degree~videos. 
	With a view selection module for finding key objects and a novel memory network leveraging two \textit{past} and \textit{future} memories, our PFMN temporally summarizes a 360\degree~video based on the underlying storyline.
}
\vspace{-10pt}
\label{fig:concept}
\end{figure}

\section{Introduction}
\label{sec:intro}

360\degree~videos are growing rapidly as recording devices such as GoPro and GearVR spread widely and many social network platforms such as YouTube and Facebook eagerly support the sharing of this content.
With the explosive growth of 360\degree~content, there is a strong need for automatic summarization, despite that 360\degree~video summarization has been still under-addressed in the video summarization literature.
In particular, only spatial summarization, which controls \textit{where} to look in the video frame of unlimited field of view (FOV) and generates an optimal camera trajectory, has been addressed in~\cite{su2016accv,su2017cvpr,hu2017cvpr}. Recently, Yu \etal~\cite{yu2018aaai} attempt to generate a spatio-temporal highlight for a long 360\degree~video, although they simply apply the ranking model for spatial summarization to the temporal domain without deliberate consideration of temporal summarization.

In this paper, we focus on story-based temporal summarization of long 360\degree~videos by selecting a subset of key subshots.
Specifically, we focus on tackling the following two summarization problems that are caused by the characteristics of 360\degree~videos.
First, 360\degree~videos contain all surroundings of a camera in all directions. 
As a result, unlike normal videos, there is no specific subject that a videographer intends to shoot. 
Since identifying the subject is often crucial to grasp the plot of the entire video,  
its absence can be a stumbling block to the temporal summarization.
Second, it is difficult to use supervised learning, which is preferable for model performance. 
There are few available pairs of long 360\degree~videos and their corresponding edited summaries. 
To make matters worse, unlimited FOVs make it difficult to browse videos, thus it takes a lot of time and effort for humans to annotate the data.


To solve aforementioned problems, we propose a novel memory network model for 360\degree~video temporal summarization, named as \textit{Past-Future Memory Network} (PFMN), whose intuition is shown in Figure~\ref{fig:concept}.
As preprocessing, we first perform the view selection using a deep ranking network that is learned from photostream data of the same topic.
That is, we compute the scores of 81 normal field of view (NFOV) region proposals cropped from the input 360\degree~video.
Then, we perform temporal summarization, with an assumption that the video set of the same topic shares the common \textit{storyline}, which we define as an ordered sequence of subshot exemplars that frequently appear in the whole video set.
We recover this latent, collective storyline, using a memory network involving
two external memories that store the embeddings of previously selected subshots and future candidate subshots that are not selected yet.
Based on the fact that humans make a video summary using the entire context, our model iteratively selects a next summary subshot using the correlation between the past and future information.
Since 360\degree~videos of good quality are not large-scale enough to train the proposed memory network, 
we pre-train the model using photostream data and fine-tune it using a small set of training 360\degree~videos.

Temporal summarization is often formulated as sequence prediction, because it requires to choose a subsequence among the entire video.
We design our model based on memory networks (\eg~\cite{graves2014arxiv,graves2016nature,gulcehre2016iclr,kumar2016icml,weston2014iclr}), instead of recurrent neural networks (RNN)~\cite{mikolov2010interspeech} and their variants such as LSTM~\cite{hochreiter1997mit} and GRU~\cite{cho2014emnlp},
which may be one of the most dominant frameworks for sequence prediction. 
We argue that the memory network approach bears two major advantages. 
First,  RNNs and their variants represent previous history with a hidden state of a fixed length, which may be often insufficient for processing a long video subshot sequence.
On the other hand, the external memory significantly improves the model's memorization capability to explicitly store the whole visual information from the video input.
Second, RNNs predict the very next item based on the hidden state only. 
However, the hidden state stores the information about the whole previous sequence, 
and some of this information, that is not relevant to the summary, may degrade the model performance.
Yet, our model stores only the previously selected subshots, which helps our model focus on the storyline of the input video.

For evaluation, we perform three sets of experiments. 
First, we run spatial summarization experiments on the Pano2Vid dataset~\cite{su2016accv}, 
to investigate the view selection capability of our model. 
Second, we evaluate story-based temporal summarization with a newly collected 360\degree~video summarization dataset,
which is the target task of this work.
Finally, we evaluate our model's summarization performance in another domain, using an image-based storytelling VIST dataset~\cite{huang2016naacl}. 
We verify that our model outperforms the state-of-the-art methods on all the tasks. 

Finally, we outline contributions of this work as follows.

\begin{enumerate}
\vspace{-3pt}\item To the best of our knowledge, our work is the first to address the problem of temporal summarization of 360\degree~videos.
\vspace{-3pt}\item We propose a novel model named \textit{Past-Future Memory Network} (PFMN).
As far as we know, our model is the first attempt to leverage memory networks for the problem of video summarization, including not only 360\degree~videos but also normal videos.
The unique updates of PFMN include (i) exploiting an additional memory for future context, and (ii) correlating the future context and previously selected subshots to select the next summary subshot at each iteration.
\vspace{-3pt}\item  We qualitatively show that our model outperforms state-of-the-art video summarization methods in various settings with
not only our own 360\degree~video dataset but also Pano2Vid~\cite{su2016accv} and VIST~\cite{huang2016naacl} benchmarks.
\end{enumerate}

\section{Related Work}
\label{sec:related_work}

\textbf{Learning Storylines.}
Our model learns an underlying visual storyline that many videos of the same topic share.
Some early methods for storyline inference require human expertise~\cite{schank2013scripts}, while 
recent approaches make use of data-driven, unsupervised learning methods~\cite{chambers2008acl,mcintyre2009acl,wang2012aaai} or weakly-supervised ones~\cite{gupta2009cvpr, xiong2015iccv}.
However, even for data-driven approaches, some pre-defined criteria like sparsity or diversity~\cite{kim2014cvpr,kim2014cvpr2} are needed for recovering storylines.
One of the closest works to ours may be~\cite{sigurdsson2016eccv}, which proposes Skipping Recurrent Neural Networks (S-RNN),
to discover long-term temporal patterns in photostream data by skipping through images by adding \textit{skipping} operations to classic RNNs.
However, our model summarizes 360\degree~videos rather than photostreams, and leverages the memory structure, instead of RNNs, to fully utilize long input sequences.

As another related task, visual storytelling is to generate coherent story sentences from an image set.
Huang \etal~\cite{huang2016naacl} recently release the visual storytelling VIST dataset, and Yu \etal~\cite{yu2017emnlp} propose a model to select representative photos and then generate story sentences from those summary photos.
Since each album in the VIST has annotations to indicate which photos are used to generate stories, (\ie which photos are representative), 
we also evaluate our summarization model on the VIST to demonstrate its performance in a different domain. 

\textbf{Temporal Video Summarization.}
\cite{truong2007acmtomm} provides a thorough overview of earlier video summarization methods.
Many approaches are unsupervised methods, but still use some heuristic criteria such as importance or representativeness~\cite{hong2009sigmm,lee2012cvpr,lu2013cvpr,khosla2013cvpr,song2015cvpr}, relevance~\cite{potapov2014eccv,chu2015cvpr}, and diversity or non-redundancy~\cite{li2010wiamis,de2011vsumm,de2014speeding,zhao2014cvpr}.
There have been some supervised methods for temporal summarization, such as \cite{gong2014nips,gygli2014eccv,chao2015uai,gygli2015cvpr,zhang2016cvpr}.
Recently, a few approaches have leveraged deep neural architectures for video summarization.
For unsupervised cases, Yang \etal~\cite{yang2015cvpr} use robust recurrent auto-encoders to identify highlight segments, and Mahasseni \etal~\cite{mahasseni2017cvpr} propose a video summarization approach based on variational recurrent auto-encoders and generative adversarial networks.
For supervised cases, Zhang \etal~\cite{zhang2016eccv} integrate LSTMs with the determinantal point process for video summarization.
Deep pairwise ranking models have been also popularly used for video highlight detection such as \cite{yao2016cvpr,gygli2016cvpr}.
However, unlike ours, no previous work is based on the memory network framework. 
Our work can be categorized  as a weakly supervised approach that uses image priors for video summarization (\eg~\cite{song2015cvpr,khosla2013cvpr,kim2014cvpr2}), although we directly use raw photostreams with no summary annotation.
Moreover, our work is the first to attempt 360\degree~video temporal summarization.

\textbf{360\degree~Video Summarization.}
Despite the explosive increase of 360\degree~video content, its summarization has been still understudied, except the \textit{AutoCam} framework~\cite{su2017cvpr,su2016accv}, \textit{deep 360 pilot}~\cite{hu2017cvpr}, and \textit{Composition View Score} (CVS) model~\cite{yu2018aaai}.
AutoCam tackles the Pano2Vid problem, which takes a 360\degree~video and generates NFOV camera trajectories as if a human videographer would take with a normal NFOV camera (\ie the spatial summarization of 360\degree~videos).
The CVS framework proposes a deep ranking model for spatial summarization to select NFOV shots from each frame of a 360\degree~video, and extends the same model for the temporal domain to generate a spatio-temporal highlight video.
However, it is hard to say that the CVS framework is designed considering the issues of temporal summarization. 
Moreover, the objective of our work is different from those of previous works in that 
we produce a concise abstraction of a 360\degree~video that recovers the underlying visual storyline, rather than assessing the importance of video segments. 

\section{Datasets}
\label{sec:datasets}

We collect a dataset of 360\degree~videos for temporal summarization, along with photostream data with the same topics.
We exploit the photostream data for two purposes: 
(i) training the 360\degree~video view selection module, and (ii) pre-training our memory network model for temporal summarization.
Their key statistics are summarized in Table~\ref{tab:360video_statistics}-\ref{tab:photostream_statistics}.

\subsection{360\degree~Video Dataset}
We newly collect 360\degree~videos from \textit{YouTube} with five query terms: \textit{wedding}, \textit{parade}, \textit{rowing}, \textit{scuba diving}, and \textit{airballooning}.
All the topics have a fairly consistent narrative structure, and involve a large volume of both 360\degree~videos and photostreams.
For each topic, we download as many videos as possible, and manually filter out those that are irrelevant to the topic.
Since these videos are used for temporal summarization, we also ignore the ones that are less than five minutes long.
In total, we obtain 285 videos with a combined duration of about 92 hours.

\subsection{Photostream Dataset}
We gather collections of Flickr photostreams from the YFCC100M dataset~\cite{thomee2015arxiv} for two topics, \textit{wedding} and \textit{parade}, 
and the dataset of Kim and Xing~\cite{kim2014cvpr} for the other topics.
We select only photostreams that include more than 100 photos taken on the same day, 
and then sort each photostream based on the photos' time taken.
In total, the dataset consists of about 85K images of 9K photostreams.

\begin{table}[t]
\centering
\fontsize{8}{9.5}\selectfont
\begin{tabular}{c|ccc}
\hline
\multirow{2}{*}{Topics} & \multicolumn{3}{c}{360\degree~Videos} \\
& \# videos & total (hr) & mean (min) \\
\hline
wedding & 50 & 27.52 & 33.03 \\
parade & 82 & 33.18 & 24.27 \\
rowing & 41 & 9.25 & 13.53 \\
scuba diving & 73 & 13.45 & 11.05 \\
airballooning & 39 & 8.83 & 13.58 \\
\hline
\end{tabular}
\vspace{5pt}
\caption{Statistics on our new 360\degree~video summarization dataset.}
\label{tab:360video_statistics}
\vspace{10pt}
\begin{tabular}{c|ccc}
\hline
\multirow{2}{*}{Topics} & \multicolumn{3}{c}{Photostreams (PS)} \\
& \# PS & \# imgs 
& \# imgs per PS \\
\hline
wedding & 1,071 & 216,768 & 205 \\
parade & 718 & 110,789 & 154 \\
rowing & 2,019 & 166,335 & 82 \\
scuba diving & 2,735 & 176,461 & 65 \\
airballooning & 2,595 & 176,093 & 68 \\
\hline
\end{tabular}
\vspace{5pt}
\caption{Statistics on our photostream dataset.}
\vspace{-13pt}
\label{tab:photostream_statistics}
\end{table}

\subsection{Video Formats}

A 360\degree~video frame cannot be projected in a 2D plane with no distortion.
Thus, we first select a viewpoint in a form of longitude and latitude coordinates $(\phi, \theta)$ in the spherical coordinate system,
and extract an NFOV region from the 360\degree~frame by a rectilinear projection with the viewpoint as center.
The ideal size of an NFOV region may need to vary according to several factors (\eg the sizes of key objects or the geometry of filming environment).
We heuristically set the size of an NFOV region to span 54\degree~horizontally and 30\degree~vertically, while fixing the aspect ratio to 16:9, because it performs the best in experiments.

If some distortion is allowed, a whole spherical frame can be represented by equirectangular projection (ERP), 
which transforms the spherical coordinates of a 360\degree~frame into the planar coordinates as a 2D world map from the spherical Earth.
The resulting $x, y$ coordinates in the ERP format are: $x = (\phi - \phi_0)\cos\theta_1, y = (\theta - \theta_1)$, where $\phi_0$ is the central meridian and $\theta_1$ are the standard parallels (north and south of the equator) in the spherical coordinate system. 

\begin{figure*}
	\begin{center}
		\includegraphics[width=1.0\textwidth]{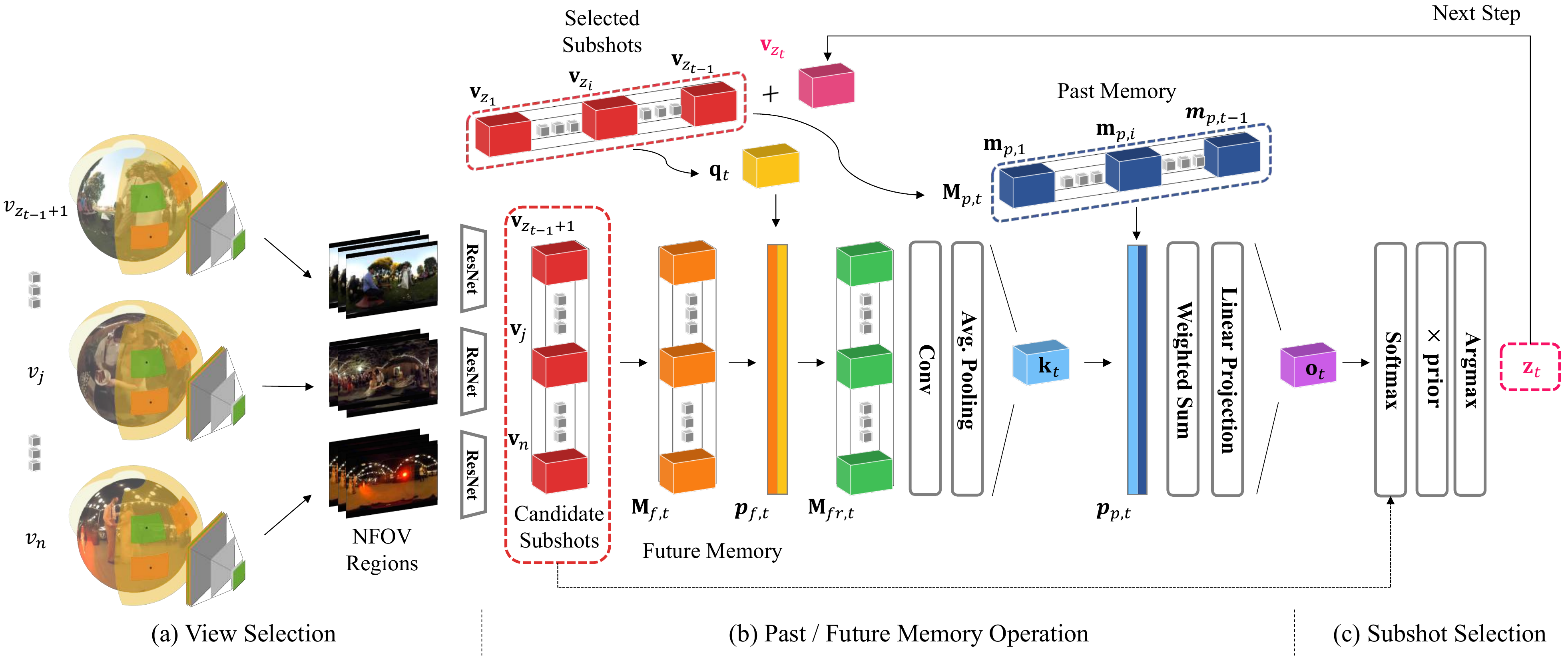}
	\end{center}
	\vspace{-10pt}
	\caption{Architecture of the proposed \textit{Past-Future Memory Network}.
		 (a) We compute the key-objectness scores of NFOV regions from each spherical subshot using a deep ranking network (section~\ref{sec:view_selection}).
		 (b) The past memory $\mathbf M_p$, which stores already selected subshots, and the future memory $\mathbf M_f$, which saves the remaining subshots, are correlated via convolutional operations and soft attention models to compute output $\mathbf o_t$ (section~\ref{sec:temp_summary}).
		 (c) Finally, the subshot from the future memory with the highest score is chosen as the next subshot summary, and move into the past memory for next iteration (section~\ref{sec:temp_summary}).
	}
	\vspace{-10pt}
	\label{fig:PFMN}
\end{figure*}


\section{Past-Future Memory Network (PFMN)}
\label{sec:pfmn}

Figure~\ref{fig:PFMN} shows the overall architecture of \textit{Past-Future Memory Network} (PFMN).
The input to the model is a sequence of video subshots $\mathcal{V} = \{v_1, v_2, \cdots, v_n\}$ for an input 360\degree~video sampled at 5 fps.
We construct $\mathcal{V}$ using the Kernel Temporal Segmentation (KTS)~\cite{potapov2014eccv}, maintaining visual coherence within each subshot.
Subshots have an average duration of 6.1s.
The output is a selected subsequence $\mathcal{S} = \{v_{z_1}, v_{z_2}, \cdots, v_{z_m}\} \subset \mathcal V$ as a summary, where $m$ is a user parameter to set the length of the summary.

We first run a preprocessing step that extracts a set of 81 NFOV candidates with key-objectness scores from each 360\degree~subshot $v_i$ (section \ref{sec:view_selection}). 
We then run temporal summarization using our proposed PFMN model (section \ref{sec:temp_summary}).
We design the view selection module by a simple modification of standard deep ranking models, and our major technical novelties lie in the memory network part of PFMN.

\subsection{360\degree~Video View Selection}
\label{sec:view_selection}

\begin{figure}[t!]
\centering
\includegraphics[trim=0.2cm 0.2cm 0cm 0.1cm,clip,width=0.47\textwidth]{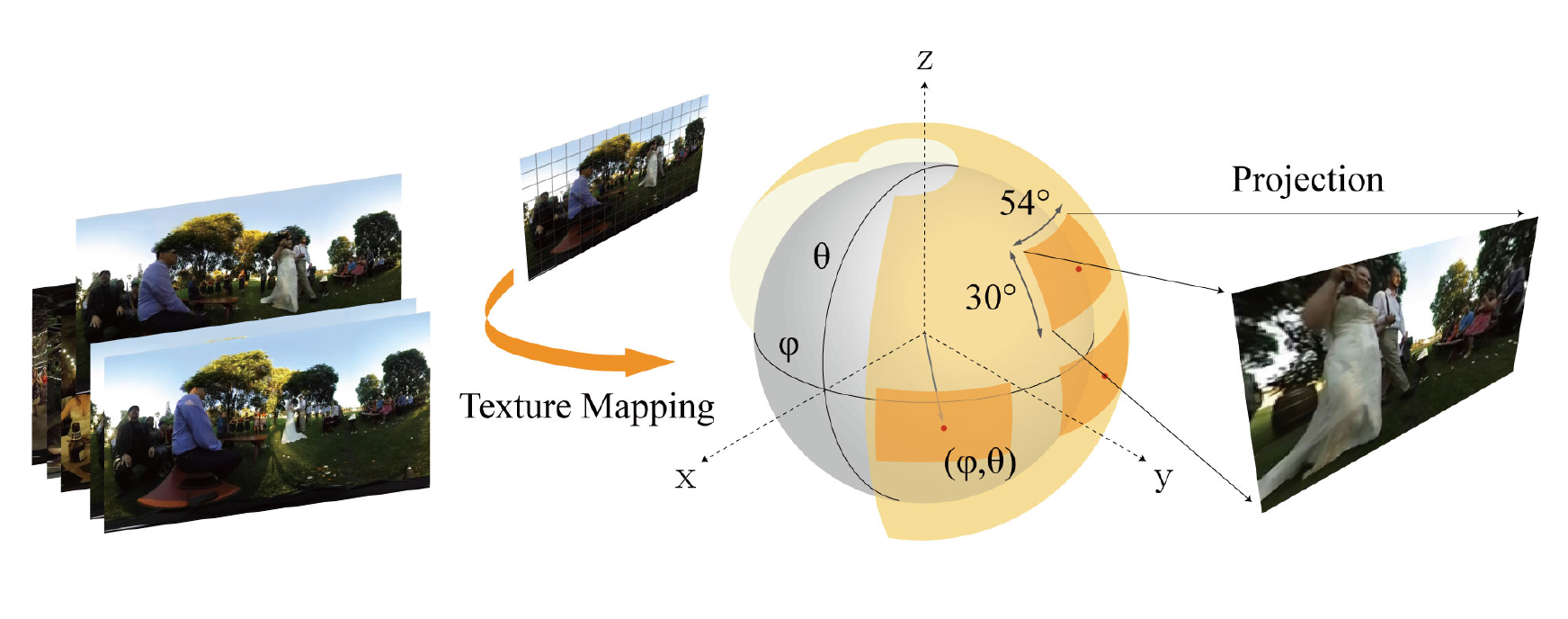}
\caption{NFOV projection.
	We extract an NFOV region from a 360\degree~frame by a rectilinear projection with a specified viewpoint as center.
	The size of each NFOV region spans a horizontal angle 54\degree~and a vertical angle 30\degree~for a 16:9 aspect ratio.}
\vspace{-15pt}
\label{fig:NFOV_projection}
\end{figure}

The goal of the 360\degree~video view selection is to assign key objectness scores to a set of NFOV  candidates from each subshot $v_i$ of a whole spherical frame.
We sample 81 NFOV candidate regions at longitudes $\phi \in \Phi = \{0\degree, 40\degree, 80\degree, \cdots, 320\degree\}$ and latitudes $\theta \in \Theta = \{0\degree, \pm15\degree, \pm35\degree, \pm55\degree, \pm75\degree\}$ from the middle frame of $v_i$, as depicted in Figure~\ref{fig:NFOV_projection}.
We use a trained deep ranking model for computing the key-objectness score of each candidate region $v_{i, j}$ for $j=1,2,\cdots,81$. 
Specifically, we define the key object as an object that people are likely to keep track of when they shoot normal videos of the same topic: for example, the bride and groom in wedding videos and tropical fishes in scuba diving videos.
Since 360\degree~videos record all surroundings of a camera in all directions, 
we want to filter out irrelevant objects and only keep the key objects as the input to a 360\degree~video summarization model. 

\begin{figure}[t!]
\centering
\includegraphics[trim=0.2cm 0.2cm 0cm 0.1cm,clip,width=0.47\textwidth]{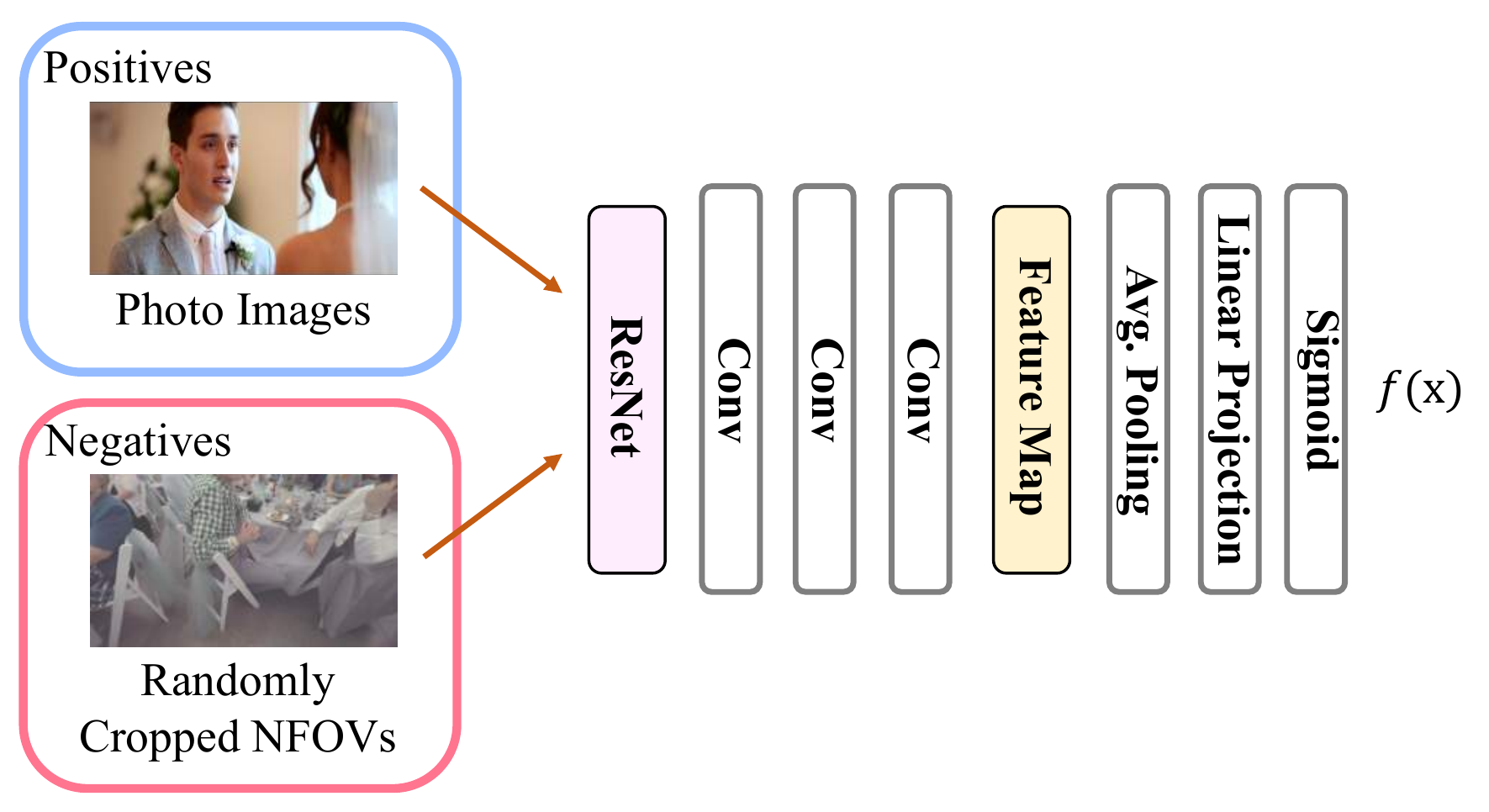}
\caption{The architecture of our view selection model. 
Unlike RankNet~\cite{gygli2016cvpr}, our ranking model consists of convolutional layers followed by a global average-pooling layer, a linear projection layer and a sigmoid activation.
We use stride 1 and no zero-padding for each convolutional layer.}
\vspace{-10pt}
\label{fig:ranking_model}
\end{figure}

\textbf{Model.}
Inspired by RankNet~\cite{gygli2016cvpr}, we leverage a deep ranking network for computing the key-objectness score $f(\mathbf x)$ of an  input NFOV $\mathbf x$ (See Figure~\ref{fig:ranking_model}).
While RankNet consists of fully-connected layers, our ranking model is based on convolutional (Conv) layers to exploit the spatial attention.
We represent the input using the res5c feature map $\mathbf x^{r5c} \in \mathbb{R}^{7 \times 7 \times 2048}$ of ResNet-101~\cite{he2016cvpr}, pretrained on ImageNet~\cite{deng2009cvpr}. We then feed it into three Conv layers with filters $\mathbf w_{rank}^{i} \in \mathbb{R}^{2 \times 2 \times c_{rank}^{i-1} \times c_{rank}^{i}}$ for $i = 1, 2, 3$:
\begin{equation} 
\label{eq:rank_conv}
\mathbf x_{rank}^{i} = \mbox{ReLU}( \mbox{conv} ( \mathbf x_{rank}^{i-1}, \mathbf{w}_{rank}^{i})),
\end{equation}
\noindent where conv (input, filter) indicates the Conv layer, and $\mbox{ReLU}$ is an element-wise ReLU activation~\cite{nair2010icml}.
We set $\mathbf{x}_{rank}^{0} = \mathbf x^{r5c}$ and $c_{rank}^{i=\{1,2,3\}} = \{2048, 1024, 512 \}$ as  filter channels.
We apply Batch Normalization~\cite{ioffe2015icml} to every Conv layer.
Finally, we apply a global average pooling, a linear projection, and a sigmoid activation to $\mathbf{x}_{rank}^{3} \in \mathbb{R}^{5 \times 5 \times 512}$, to compute key-objectness score $f(\mathbf x)$.

\textbf{Training.}
We regard photo images as positives, and randomly cropped NFOV regions as negatives.
That is, we define the ranking constraint over the training set: 
$ f(p_i) \succ f(c_i), \forall~(p_i, c_i) \in \mathcal{D}_{rank}$,
where $p_i$ and $n_i$ are positive and negative samples, respectively.
We then train our ranking model using the max-margin loss $\mathcal{L}_{rank,i} = \max(0, f(n_i) - f(p_i) + 1)$. 
The final objective is the total loss over $\mathcal{D}_{rank}$ with $l_2$ regularization: 
\begin{equation} 
\label{eq:max_margin}
\mathcal{L}_{rank} = \sum_{i} \mathcal{L}_{rank,i} + \lambda||\mathcal{M}_{rank}||_{F}^2,
\end{equation}
\noindent where $\mathcal{M}_{rank}$ denotes the model parameters and $\lambda$ is a regularization hyperparameter.

\subsection{Story-based Temporal Summarization}
\label{sec:temp_summary}

When humans summarize a video, they may first watch the whole video to understand the entire context, 
and then choose one by one by comparing between what they have already selected and what remains in the video. 
Based on this intuition, 
our temporal summarization includes two external memories: \emph{past memory} and \emph{future memory}, 
as one of our key novelties. 
The past memory $\mathbf M_p$ stores the already selected subshots as summary, while the future memory $\mathbf M_f$ stores the remaining subshots of the video. 
Initially, the past memory is empty (\ie $\mathbf M_p^{(0)} = \emptyset$), and the future memory includes all the subshots of the video (\ie $\mathbf M_f^{(0)} = \mathcal V$). 
At each iteration $t$, we select $v_{z_t}$ from the future memory as the best (\ie the most plausible) subshot to be included in the summary, and then move $v_{z_t}$ to the past memory. At next iteration $t+1$, the future memory becomes $\mathbf M_f^{(t+1)} = \{v_{z_{t}+1}, \cdots, v_{n}\}$; we only allow the subshots after the latest summary subshot as candidates to be chosen.
We iterate this process $m$ times; eventually the summary is identical to the past memory elements: $\mathcal S = \{v_{z_1}, v_{z_2}, \cdots, v_{z_m}\} = \mathbf M_p^{(m)}$.
We assume here that the summary subshots are sorted in a chronological order.

\textbf{Subshot Representation.}
We convert each subshot $v_i$ into feature representation as follows. 
For each of 81 NFOV candidates $\{v_{i, j}\}_{j=1}^{81}$ from $v_i$, 
we extract the pool5 feature vector $\mathbf v_{i, j}$ of ResNet-101~\cite{he2016cvpr} pre-trained on ImageNet.
We then sum over all regions as $\mathbf v_{i} = \sum_{j} w_j \mathbf v_{i,j} \in \mathbb{R}^{2,048}$, where $w_j$ is the normalized key-objectness score of $v_{i,j}$ that the view selection model computes.

\textbf{Past and Future Memory.}
At iteration $t$, the past memory stores previous summary subshots $\{v_{z_1}, \cdots, v_{z_{t-1}}\}$.
Following the standard representation of memory slots (\eg\cite{sukhbaatar2015nips}),
we encode each of them into input and output embeddings using different parameters $\mathbf{W}_{p}^{i/o} \in \mathbb{R}^{1024\times2048}$ and $\mathbf{b}_{p}^{i/o} \in \mathbb{R}^{1024}$, 
where $i$ and $o$ stand for input and output embeddings, respectively.
The resulting past memory vector $\mathbf{m}_{p,j}^{i/o} \in \mathbb{R}^{1024}$ is represented by
\begin{align}
\label{eq:p_memory}
\mathbf{m}_{p,j}^{i/o} = \mbox{ReLU}(\mathbf{W}_{p}^{i/o} \mathbf v_{z_j} + \mathbf{b}_{p}^{i/o}), \hspace{6pt} j = 1,\cdots,z_{t-1}.
\end{align}

The future memory stores the information about future subshots, $\{ v_{z_{t-1}+1}, \cdots, v_n \}$.
Similar to the past memory, the future memory vector $\mathbf{m}_{f, j}^{i/o} \in \mathbb{R}^{1024}$ is represented as input and output embeddings:
\begin{equation}
\label{eq:f_memory}
\mathbf{m}_{f, j}^{i/o} = \mbox{ReLU}(\mathbf{W}_{f}^{i/o} \mathbf v_{j} + \mathbf{b}_{f}^{i/o}), \hspace{6pt} j = z_{t-1}\hspace{-1pt} +\hspace{-1pt} 1, \cdots \hspace{-1pt}, n
\end{equation}
\noindent where $\mathbf{W}_{f}^{i/o} \in \mathbb{R}^{1024\times{2048}}$ and $\mathbf{b}_{f}^{i/o} \in \mathbb{R}^{1024}$.

Finally, we stack the memory vectors row by row for later computation:
$\mathbf{M}_{p, t}^{i/o} = [(\mathbf{m}_{p, 1}^{i/o})^{\mathsmaller T};\cdots;(\mathbf{m}_{p, t-1}^{i/o})^{\mathsmaller T}]$ and $\mathbf{M}_{f, t}^{i/o} = [(\mathbf{m}_{f, z_{t-1}+1}^{i/o})^{\mathsmaller T};\cdots;(\mathbf{m}_{f, n}^{i/o})^{\mathsmaller T}] $. 

\textbf{Correlating between Past and Future Memory.}
Rather than using recurrent connections throughout time as RNNs do, our model predicts the next summary subshot by extracting a key vector from the future memory, and computing the attention of the past memory with respect to the key vector.
Inspired by RWMN~\cite{na2017iccv}, we use convolutions with a learnable kernel as a read operation on the memory.
We compute a query embedding $\mathbf{q}_t$ from  the mean-pooled summary $\mathbf{v}_{avg} = (\sum_{j=1}^{t-1} \mathbf v_{z_j}) / (t-1)$:
\begin{equation}
\label{eq:q_t}
\mathbf{q}_t = \mbox{ReLU}(\mathbf{W}_{q} \mathbf{v}_{avg} + \mathbf{b}_{q}),
\end{equation}
\noindent
where $\mathbf{W}_{q} \in \mathbb{R}^{1024 \times 2048}$ and $\mathbf{b}_{q} \in \mathbb{R}^{1024}$.
$\mathbf{q}_t$ can be interpreted as an indicator of the current story flow based on the previously selected summary subshots $\{v_{z_1}, \cdots, v_{z_{t-1}}\}$. 
We intend to dynamically update the future memory embedding according to $\mathbf{q}_t$ at every step.
Based on this intuition, $\mathbf{q}_{t}$ is fed into the soft attention model of the future memory:
\begin{align}
\label{eq:p_f,t}
\mathbf{p}_{f,t} &= \mbox{softmax}(\mathbf{M}_{f, t}^{i} \mathbf{q}_{t}), \\
\label{eq:M_fr,t}
\mathbf{M}_{fr, t}[i,:] &= \mathbf{p}_{f,t}[i] \mathbf{M}_{f,t}[i,:],
\end{align}
\noindent
where $\mathbf{p}_{f, t} \in \mathbb{R}^{n-z_{t-1}}$ and $\mathbf{M}_{fr, t} \in \mathbb{R}^{(n-z_{t-1})\times1024}$.
It means that we compute how well the query embedding $\mathbf{q}_{t}$ is compatible with each cell of future memory by softmax on the inner product (Eq.(\ref{eq:p_f,t})), 
and rescale each cell by the element-wise multiplication with the attention vector $\mathbf{p}_{f,t}$ (Eq.(\ref{eq:M_fr,t})).
We regard the attended output memory $\mathbf{M}_{fr, t}$ as the summarization context.

Since adjacent subshots in videos connecting storylines often have strong correlations with one another, we associate neighboring memory cells of the attended memory by applying a Conv layer to $\mathbf{M}_{fr, t}$.
The Conv layer consists of a filter $\mathbf{w}_{fr} \in \mathbb{R}^{k_v \times k_h \times 1024 \times 1024}$, whose vertical and horizontal filter size and strides are $k_v = 20, k_h = 1024, s_v = 10, s_h = 1$, respectively.
We then apply an average-over-time pooling to obtain the key vector $\mathbf{k}_{t} \in \mathbb{R}^{1024}$, which can be regarded as a concise abstraction of future context with respect to the current query $\mathbf{q}_{t}$:
\begin{equation}
\label{eq:k_t}
\mathbf{k}_{t} = \mbox{averagepool}(\mbox{conv}(\mathbf{M}_{fr, t}, \mathbf{w}_{fr}, \mathbf{b}_{fr})),
\end{equation}
\noindent
where $\mathbf{b}_{fr} \in \mathbb{R}^{1024}$ is a bias vector.

We then compute the soft attention of the past memory through the inner product between $\mathbf{k}_{t}$ and each cell of $\mathbf{M}_{p,t}^{i}$. The final memory output $\mathbf{m}_{t} \in \mathbb{R}^{1024}$ becomes an abstraction of the past memory $\mathbf{M}_{p,t}$ using the future context $\mathbf{k}_t$:
\begin{align}
\label{eq:m_t}
\mathbf{m}_{t} = \mathbf{p}_{p,t}^{\mathsmaller T}\mathbf{M}_{p,t}^{o}, \hspace{6pt}
\mathbf{p}_{p,t} = \mbox{softmax}(\mathbf{M}_{p, t}^{i} \mathbf{k}_{t}), 
\end{align}
\noindent
where $\mathbf{p}_{p,t} \in \mathbb{R}^{t-1}$.
This part is the contact point where past and future information are combined, and is the primary novelty of our model.

\textbf{Selecting the Next Summary.}
Finally, we select the next summary using the final memory output $\mathbf{m}_{t}$.
By using a linear projection, we extend the dimension of $\mathbf{m}_{t}$ to the output $\mathbf{o}_t \in \mathbb{R}^{2048}$,
and compute the compatibility scores $c_j$ between $\mathbf{o}_{t}$ and each future subshot $\mathbf{v}_j$ for $j=z_{t-1}+1, \cdots, n$, using the inner product and the softmax operation: $c_j = \mbox{softmax}(\mathbf{o}_t^{'}\mathbf{v}_j)$.

We then multiply $c_j$ by a prior $u_{j,t}$ to be the selection probability $s_j = c_j u_{j,t}$.
The prior is the probabilistic equivalent of random sampling without replacement, considering the selected subshot sequence is an ordered subset:
\begin{align}
u_{j,t} = \left\{ \begin{array}{ll}
\prod_{k=z_{t-1}+1}^{j-1}(1-u_{k,t})  \frac{m-t+1}{n-t+1}  & j \le n-m+t, \\
0 & j > n-m+t.\\
\end{array} \right. \nonumber
\end{align}
\noindent
This prior simply conveys two obvious constraints: (i) the farther away from the subshot selected in the previous iteration, the lower the probability is assigned, and (ii) the probability of a subshot that should not be selected at the current iteration is zero (\eg when $n=10, m=4$, it prevents selecting the ninth subshot as the second summary).

Finally, we select the next summary $v_{z_t}$ in a greedy way; that is, we select the subshot with the highest selection probability: $z_t = \argmax_{j \in \{z_{t-1}+1, \cdots, n\}} s_j$.
Until the number of selected subshots reaches a  user-defined $m$, the selected summary $v_{z_t}$ moves back into the past memory, and we repeat all the above selection process again.

\subsection{Training}
\label{sec:training} 

The goal of training is to learn transitions between summary subshots, so as to recover the latent storyline and summarize individual videos.
Since we have no annotation for true key subshots in the training set, we optimize the likelihood of a selected subshot at each iteration $t$ (\ie maximize the softmax probability $c_{z_{t}}$).
This is mathematically based on the training procedure of S-RNN~\cite{sigurdsson2016eccv}, which ensures that the model parameters converge to a local optimum.
Since 360\degree~video training data are not large-scale, we use both photostream data and 360\degree~video data for training.
We first train the parameters of the memory network using photostreams, and then fine-tune using the 360\degree~video training set. 
Note that our method does not require groundtruth summary for both photostreams and 360\degree~videos, and instead uses the raw photostreams and videos for training. 

Thus, our loss function over the training set $\mathcal{D}$ is
\begin{equation} 
\label{eq:final_loss}
\mathcal{L} = \sum_{\mathcal{V} \in \mathcal{D}} \sum_{j=1}^{m} -\log(c_{z_{j}}),
\end{equation}
where $\mathcal{V}$ is a photostream for pre-training and a 360\degree~video for fine-tuning.
To make the view selection more optimized for summarization, 
we fine-tune the parameters of the view selection model along with those of the memory network.

\textbf{Implementation Details.}
We initialize all the parameters via random sampling from a Gaussian distribution with standard deviation of $\sqrt{2/\textrm{dim}}$, following He \etal~\cite{he2015arxiv}.
For optimizing the objective of the view selection in Eq.(\ref{eq:max_margin}), we use a SGD with a mini-batch size of 16, a Nesterov momentum of 0.5, and $\lambda$ to $1e-07$. 
We set our initial learning rate as 0.0001 and divide it by 2 at every 16 epochs.
For optimizing the objective of the memory network in Eq.(\ref{eq:final_loss}), we select the AdaGrad~\cite{duchi2011jmlr} optimizer with a a learning rate of 0.001 and an initial accumulator value of 0.1.

\section{Experiments}
\label{sec:experiments}

We evaluate the PFMN from three aspects. 
First, we run spatial summarization experiments on the Pano2Vid dataset~\cite{su2016accv}, 
to study the view selection of our model. 
Second, we evaluate the story-based temporal summarization on our newly collected 360\degree~video dataset, on which
our model achieves the state-of-the-art performance. 
Finally, we evaluate our model's performance in another domain; 
using the image-based storytelling dataset (VIST)~\cite{huang2016naacl}. 
\vspace{-3pt}

\subsection{Experimental Setting}
\label{sec:experiment_setting}
\textbf{View Selection.}
As a proxy study for capturing key objects, we measure the performance of spatial summarization on 
the Pano2Vid dataset, consisting of 86 360\degree~videos of four topics. 
The center coordinates of the selected regions are annotated in the spherical coordinates (\ie latitude and longitude) at each segment.
Using them as groundtruth (GT), we compare the similarity between the human-selected trajectories and algorithm-generated ones.
We use the metrics of mean cosine similarity and mean overlap as in the Pano2Vid benchmark~\cite{su2016accv}.

\textbf{Temporal Summarization.}
For evaluation, we randomly sample 10 360\degree~videos per topic as a test set, and then obtain three GT summaries per video from human annotators. 
To avoid the chronological bias~\cite{song2015cvpr}, a tendency that humans prefer the shots that appear earlier in videos,
we construct the GTs of key subshots as follows.
For each test video, we ask an annotator to watch the whole video and mark all the frames that they think are important in the story flow.
We then segment the video into subshots using the Kernel Temporal Segmentation (KTS)~\cite{potapov2014eccv}, 
and rank the subshots in the descending order by $f_{is}/f_i$, where $f_i$ is the number of frames in subshot $i$, and $f_{is}$ is the number of selected frames. 
For annotation, we let the total duration of GT to be below 15\% of the whole video.
We recruit five annotators with different backgrounds. 
Following video summarization literature, we use the average pairwise $\textrm{F}_{1}$-measure compared to GT as evaluation metrics.

\textbf{Storyline Evaluation.}
The VIST dataset consists of 10,000 Flickr photo albums with 200,000 images for 69 topics.
Each album includes 10 to 50 photos taken within a 48-hour span with two GT summaries, each consisting of five human-selected photos.
We select twelve topics with the largest number of albums. 
We set aside 10\% of the training set for validation. 
We compute the precision and recall of generated summaries using the combined set of GT stories. 

\newcolumntype{P}[1]{>{\centering\arraybackslash}p{#1}}
\newcolumntype{L}[1]{>{\arraybackslash}p{#1}}
\begin{table}[t!]
\centering
\fontsize{8}{9.5}\selectfont
		\begin{tabular}{l|cc}
			\hline
			
			\multirow{2}{*}{Methods}        & Frame & Frame \\   
			& cosine sim & overlap            \\ \hline
			Center~\cite{su2016accv}                  & 0.572                   & 0.336              \\
			Eye-Level~\cite{su2016accv}               & 0.575                   & 0.392              \\
			Saliency~\cite{su2016accv}                & 0.387                   & 0.188              \\ \hline
			AutoCam~\cite{su2016accv}                 & 0.541                   & 0.354              \\   
			AutoCam-stitch~\cite{su2016accv}          & 0.581                   & 0.389              \\
			TS-DCNN~\cite{yao2016cvpr}                & 0.578                   & 0.441              \\    
			RankNet~\cite{gygli2016cvpr}              & 0.562                   & 0.398              \\ \hline 
			PFMN                                      & \textbf{0.661}          & \textbf{0.536}      \\ \hline
			
		\end{tabular}
		\vspace{5pt}
		\caption{
			Experimental results of spatial summarization on the Pano2Vid~\cite{su2016accv} dataset. Higher values are better in both metrics.
		}
		\vspace{-10pt}
		\label{tab:results_view}
\end{table}

\begin{table}[t!]
\centering
\fontsize{8}{9.5}\selectfont
		\begin{tabular}{l|c}
			\hline
			Methods        			 			& $\textrm{F}_{1}$-measure (\%)  \\ \hline   
			random sampling					  	& 13.01 \\
			uniform sampling     				& 14.60 \\
			VSUMM~\cite{de2011vsumm,de2014speeding}			& 12.38 \\ 
			SUM-GAN~\cite{mahasseni2017cvpr}	& 17.72 \\ \hline
			TS-DCNN~\cite{yao2016cvpr}			& 16.24 \\
			RankNet~\cite{gygli2016cvpr}		& 15.96 \\ 
			S-RNN~\cite{sigurdsson2016eccv}     & 19.59 \\ \hline
			PFMN-FF								& 23.15 \\ 
			PFMN-FA								& 24.40 \\
			PFMN-PA								& 24.14 \\ \hline
			PFMN-noview                         & 23.28 \\
			PFMN-RankNet						& 23.69 \\
			PFMN-hard							& 24.27 \\ \hline
			PFMN								& \textbf{24.60} \\ \hline
			
		\end{tabular}
		\vspace{5pt}
		\caption{
            Experimental results of temporal summarization on our 360\degree~video dataset.
		}
		\vspace{-6pt}
		\label{tab:results_temporal}
\end{table}

\begin{table}[t!]
\centering
\fontsize{8}{9.5}\selectfont
		\begin{tabular}{l|cc}
			\hline
			Methods        			 			& Precision (\%) & Recall (\%) \\ \hline   
			K-means                  			& 43.76     	 & 27.71  \\
			S-RNN~\cite{sigurdsson2016eccv}     & 45.17     	 & 28.57  \\
			S-RNN-~\cite{sigurdsson2016eccv}    & 39.91     	 & 25.22  \\ 
			h-attn-rank~\cite{yu2017emnlp}      & 45.30     	 & 28.90 \\ \hline   
			PFMN                                & \textbf{50.42} & \textbf{31.85} \\ \hline
			
		\end{tabular}
		\vspace{5pt}
		\caption{
			Experimental results of storyline evaluation on the image-based (VIST)~\cite{huang2016naacl} dataset.
		}
		\vspace{-13pt}
		\label{tab:results_story}
\end{table}

\begin{figure*}
	\begin{center}
		\includegraphics[width=0.99\textwidth]{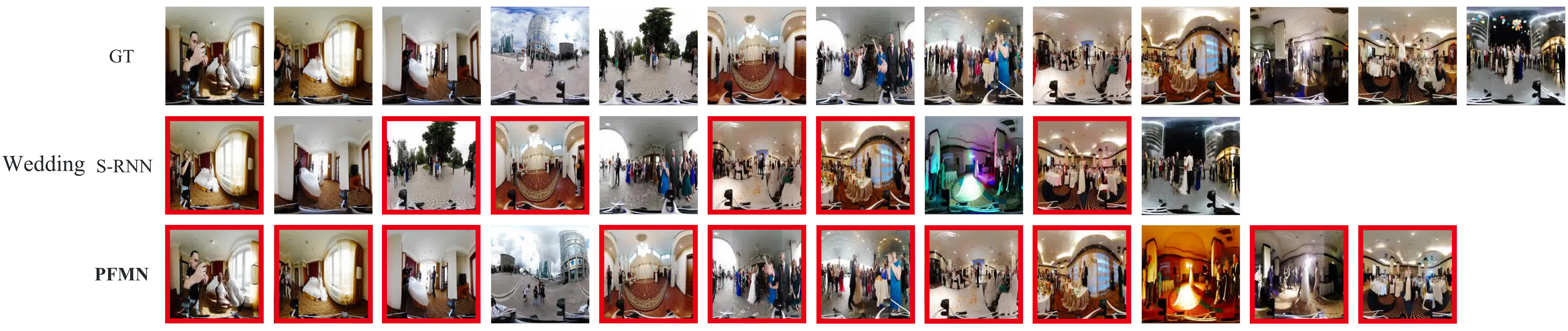}
	\end{center}
	\vspace{-6pt}
	\caption{Qualitative examples of 360\degree~video summaries generated by S-RNN~\cite{sigurdsson2016eccv} and our PFMN, along with groundtruth (GT) summaries.
		Each subshot is represented by a sampled frame in the ERP format. Red boxes indicate the matches with the GT and prediction.
	}
	\vspace{-5pt}
	\label{fig:examples}
\end{figure*}

\subsection{Baselines}
\label{sec:baselines}
For view selection (\ie spatial summarization), we use six baselines: (i) three simple baselines of spatial summarization (Center, Eye-level and Saliency) used in~\cite{su2016accv}, (ii) AutoCam~\cite{su2016accv}, and (iii) two state-of-the-art pairwise deep ranking models for highlight detection, TS-DCNN~\cite{yao2016cvpr} and RankNet~\cite{gygli2016cvpr}.
Since deep 360 pilot~\cite{hu2017cvpr} requires labeled data for training, it is not selected as our baseline.

For temporal summarization, we use five unsupervised algorithms (random/uniform sampling, VSUMM~\cite{de2011vsumm,de2014speeding}, SUM-GAN~\cite{mahasseni2017cvpr}, and S-RNN~\cite{sigurdsson2016eccv}) and two deep ranking algorithms (TS-DCNN and RankNet).
The random/uniform sampling, VSUMM, and SUM-GAN are keyframe based methods. 
Thus, we transform the automatically selected frames into key subshots as done in constructing subshot GTs in section~\ref{sec:experiment_setting}.
For TS-DCNN and RankNet, we use photostream and randomly selected video subshots as positive and negative samples, respectively.

As an ablation study, we test six variants of our method. 
First, we use three different configurations of past and future memory: (i) the future memory with only 5\% of the remaining subshots (PFMN-FF) or (ii) the whole subshots in the entire video (PFMN-FA), and (iii) the past memory with all the subshots that are not in the future memory (PFMN-PA).
Furthermore, we test three variants with different view selection methods: (iv) with no view selection (PFMN-noview), (v) view selection with RankNet (PFMN-RankNet) and (vi) hard view selection (PFMN-hard) that chooses  only top-$K$ NFOV regions. 
We set $K$ to 12.

For storyline evaluation, we use four baselines: K-means clustering, S-RNN, S-RNN- and a hierarchically-attentive RNN with ranking regularization (h-attn-rank)~\cite{yu2017emnlp}.

We implement the baselines as follows. For h-attn-rank and VSUMM, we use the code by the authors. 
For AutoCam and their baselines (Center, Eye-level and Saliency), we simply report the numbers in the original paper. 
We implement the other baselines using PyTorch~\cite{paszke2017nips} by ourselves.

\subsection{Quantitative Results}
\label{sec:quanti_results}

\textbf{View Selection.}
Table~\ref{tab:results_view} compares the performance of our PFMN with those of baselines for spatial summarization on the Pano2Vid dataset.
PFMN outperforms all the baselines by substantial margins in terms of both frame cosine similarity and frame overlap. 
It implies that our model captures key objects well in 360\degree~views.
Except the effect of the smooth-motion constraint (AutoCam-stitch), which enforces that longitude and latitude differences between consecutive segments must be less or equal than 30\degree~so as to produce better camera trajectories ($|\phi_t - \phi_{t-1}|, |\theta_t - \theta_{t-1}| \le 30\degree$), all the ranking models (RankNet, TS-DCNN and PFMN) achieve better results than the binary classification model (AutoCam).
This suggests that formulating the spatial summarization as a ranking problem is more appropriate than as a binary classification problem of whether a view is in good composition or not.
With the smoothing constraint, our PFMN even outperforms the improved version of AutoCam~\cite{su2017cvpr}, which allows three FOVs per segment for view selection rather than a single fixed FOV (PFMN: 0.641 vs.~\cite{su2017cvpr}: 0.630 in terms of the frame overlap).

\textbf{Temporal Summarization.}
Table~\ref{tab:results_temporal} shows the results of temporal summarization on our new 360\degree~video dataset.
Our PFMN achieves better results compared to baselines with large margins.
Among our variants, the PFMN with only the next few subshots leads to a performance drop (PFMN-FF: -1.45). 
The performance also slightly decreases when we use the whole video subshots for the future memory (PFMN-FA: -0.20).
This may be due to the information redundancy with the past memory.
Too much information in the past memory also drops the performance (PFMN-PA: -0.46), which implies that unnecessary subshots in the past memory may distract the selection of the next-best summary subshot.
We also validate that the quality of view selection is critical for the performance of 360\degree~video summarization (PFMN-noview: -1.32, PFMN-RankNet: -0.91).
The small gap between hard and soft view selection (PFMN-hard: -0.33) indicates that only a small region of each 360\degree~frame contains important objects.


\textbf{Storyline Evaluation.}
Table~\ref{tab:results_story} shows the results of storyline evaluation on the image-based VIST dataset.
For fair comparison, we use the same feature representation, the pool5 feature of ResNet-101~\cite{he2016cvpr}, for all models.
Our PFMN has higher performance compared to all the baselines, suggesting that our model also learns the latent storylines from images more successfully than baseline models, although our method is designed for 360\degree~videos.
It is notable that our model even achieves better results than h-attn-rank, which is a supervised method.

\subsection{Qualitative Results}
\label{sec:quali_results}

Figure~\ref{fig:examples} shows some qualitative examples of temporal summarization on 360\degree~test videos of our dataset.
In 360\degree~subshots in the ERP (equirectangular projection) format, we show the matched subshots between the prediction and GT summaries in red boxes.
The proposed PFMN not only recovers much of the GT summaries labeled by human annotators, but also successfully captures the main events in the storyline of the test video.

\section{Conclusion}
\label{sec:conclusion}

We proposed the \textit{Past-Future Memory Network} model for story-based temporal summarization of 360\degree~videos.
Our model recovered latent, collective storyline using a memory network that involves two external memories to store the embeddings of previously selected subshots and future candidate subshots. 
We empirically validated that the proposed memory network approach outperformed other state-of-the-art methods,
not only for view selection but also for story-based temporal summarization in both 360\degree~videos and photostreams. 
We believe that there are several future research directions that go beyond this work. 
First, we can apply our approach to other types of summarization in the domains of text or images, because 
the proposed memory network has no limitation on the data modality. 
Second, we can extend our method to be a purely unsupervised method that automatically detects the storylines from only 360\degree~video corpus without aid of photostreams.

\textbf{Acknowledgements}.
This work was supported by the Visual Display Business (RAK0117ZZ-21RF) of Samsung Electronics, 
and IITP grant funded by the Korea government (MSIT) (No. 2017-0-01772).
Gunhee Kim is the corresponding author.

{\small
\bibliographystyle{ieee}
\bibliography{ms}
}

\end{document}